%% file: main.tex
\documentclass[10pt,twocolumn,letterpaper]{article}

\usepackage[pagenumbers]{cvpr}      
\usepackage[normalem]{ulem}
\usepackage{booktabs}
\usepackage{multirow}
\usepackage{siunitx}

\input{preamble}

\definecolor{cvprblue}{rgb}{0.21,0.49,0.74}
\usepackage[pagebackref,breaklinks,colorlinks,citecolor=cvprblue]{hyperref}

\def\paperID{6622} 
\def\confName{CVPR}
\def\confYear{2024}

\title{BiPer: Binary Neural Networks using a Periodic Function}

\author{Edwin Vargas$^{1,\dagger}$, Claudia V. Correa$^{1}$, Carlos Hinojosa$^{2}$, Henry Arguello$^{1}$\\
$^{1}$Universidad Industrial de Santander (UIS); \ \ $^{2}$King Abdullah University of Science and Technology\\
{\tt\small $^\dagger$edwin.vargas4@correo.uis.edu.co}
}

\begin{document}
\maketitle
\input{sec/0_abstract}    
\input{sec/1_introduction}
\input{sec/2_related_works}
\input{sec/3_method}

\input{sec/4_experiments}
\input{sec/5_conclusion}
{
    \small
    \bibliographystyle{ieeenat_fullname}
    \bibliography{main}
}


\end{document}

%% file: preamble.tex
\usepackage{mathtools}
\usepackage{amsmath}
\usepackage{algorithm}
\usepackage[noend]{algpseudocode}
\usepackage{array}
\usepackage{graphicx} 
\usepackage[dvipsnames]{xcolor}
\usepackage{amssymb}
\usepackage{amsmath}
\usepackage{float}
\usepackage{mathtools}
\usepackage{algorithm}
\usepackage[noend]{algpseudocode}
\usepackage{color}
\usepackage{multirow}
\usepackage{booktabs}
\usepackage{url}
\usepackage[utf8]{inputenc}

\usepackage[accsupp]{axessibility}



%% file: sec/0_abstract.tex
\begin{abstract}
Quantized neural networks employ reduced precision representations for both weights and activations. This quantization process significantly reduces the memory requirements and computational complexity of the network. Binary Neural Networks (BNNs) are the extreme quantization case, representing values with 
just one bit. 
Since the sign function is typically used to map real values to binary values, smooth approximations are introduced to mimic the gradients during error backpropagation. Thus, the mismatch between the forward and backward models corrupts the direction of the gradient causing training inconsistency problems and performance degradation.
In contrast to current BNN approaches, we propose to employ a binary periodic (BiPer) function during binarization. Specifically, we use a square wave for the forward pass to obtain the binary values and employ the trigonometric sine function with the same period of the square wave as a differentiable surrogate during the backward pass. We demonstrate that this approach can control the quantization error by using the frequency of the periodic function and improves network performance.
Extensive experiments validate the effectiveness of BiPer in benchmark datasets and network architectures, with improvements of up to $1\%$ and $0.69\%$ with respect to state-of-the-art methods in the classification task over CIFAR-10 and ImageNet, respectively. Our code is publicly available at \url{https://github.com/edmav4/BiPer}.
\end{abstract}
\vspace{-1em}

%% file: sec/1_introduction.tex
\section{Introduction}
\label{sec:introduction}


Deep Neural Networks (DNN) have achieved unprecedented results in many high-level tasks, such as classification, segmentation, and detection, with a tremendous concurrent impact in  computer vision, natural language processing, information retrieval, and many others \cite{YoloV7}. Typically, DNNs rely on full-precision (32 bit) weights and activation functions. Accurate and precise models, however, become expensive in terms of computation, storage and number of parameters. For this reason, DNN deployment is usually prohibited for devices with limited resources, such as mobile, hand-held or wearables. 
Different approaches to reduce computation requirements include efficient neural network architecture design \cite{mobilenets,efficientnet,squeezenet}, network pruning \cite{He_2018_ECCV}, knowledge distillation \cite{distillation2021}, low rank tensor decomposition \cite{Phan_tensor_2020}, hashing \cite{HashingCompression}, and network quantization \cite{BNNSpringer_2023,BNNReview_2020,ReviewBNN2023}. Among them, network quantization has become one of the most promising techniques, aiming at compressing large models usually stored as floating-point weights with low bitwidth numbers. Binary Neural Networks (BNNs) are the extreme quantization case, where weights and activation functions are constrained to just one bit, i.e., binary values, typically +1 or -1. 
In contrast to DNNs, BNNs replace heavy matrix computations by bit-wise operations, yielding to $32\times$ memory compression, and  $58\times$ speed-up on CPUs \cite{xnornet_2016}. 
Thus, this approach drastically reduces the computational requirements and accelerates inference, making BNNs particularly appealing for resource-constrained environments such as edge devices and mobile applications.

Despite significant advantages for efficient BNN deployment in hardware with limited capabilities, the binarization of full-precision models severely degrades the accuracy performance in high-level tasks such as object detection, classification, segmentation, and others \cite{ReviewBNN2023}. For instance, in large datasets such as ImageNet, one of the earliest BNN models, the XNOR-Net \cite{xnornet_2016}, achieved an accuracy degradation of around $18\%$ compared to the full precision ResNet-18 architecture. 
Recent efforts have been devoted to close the performance gap of BNN with respect to their real-valued counterparts. Nonetheless, state-of-the-art approaches still exhibit accuracy degradations of approximately $8\%$ \cite{Recu2021}.

Binarization of real-valued weights and activations is generally performed using the sign function during the feed-forward procedure. A relevant limitation of the sign function is that its gradient is null everywhere except in zero, which makes it incompatible with error back-propagation methods, due to the non-differentiability of binary operations. To overcome this issue, various techniques like the straight-through estimator (STE) and relaxed training approaches have been adopted \cite{BNN_2016}. STE essentially substitutes the sign function for the identity function to calculate the gradients during the backwards process. Since there exists a mismatch between the forward and backward pass caused by the STE approximation, research efforts have focused on designing better smooth and differentiable functions 
to estimate the gradient of the sign function \cite{DSQ_2019,AdaSTE2022}.  
Although these approaches have improved the accuracy of BNNs, gradient instability persists when the quantization error is minimized, i.e., when the approximation functions are close to the sign function. 
Therefore, there is still a significant performance gap between real and binary models, and the pursuit of refined quantization functions for BNN, with lower accuracy degradation and higher gradient stability, remains an open challenge.


Instead of using the Sign function, in this work, we propose to address the aforementioned issues of extreme 1-bit quantization 
by using a binary periodic (BiPer) function or square wave function to promote binary weight values. Thus, opposite to the sign function 
which is always negative for negative values or positive for positive values, the proposed periodic function can reach positive and negative values in the whole domain of the latent weights. To the best of our knowledge, this is the first time a binary periodic function is employed for the binarization problem in quantized neural networks. Since the gradient of the periodic function still faces the problem of being zero almost everywhere, it cannot be directly integrated within a back-propagation algorithm based on gradient descent. We solved this problem by employing a sinusoidal function with the same fundamental frequency of the periodic function as a differentiable surrogate during the backward pass. The continuity and differentiable characteristics of the sine function, make it suitable for stochastic gradient methods. 
In contrast to existing BNN methods that smoothly and progressively approximate the sign function to reduce the quantization error (QE), we will show that in the proposed BiPer approach the QE can be controlled by 
the frequency of the periodic function. We further leverage this property to provide an initialization of the weights that better balances the 
trade-off between the estimation error and performance accuracy. Experimental results demonstrate that BiPer provides the best network performance for the classification task, with respect to state-of-the-art BNN approaches on the CIFAR-10 and ImageNet data sets.

\noindent \textbf{Paper Contribution.} The contributions of our work are summarized as follows:
\begin{itemize}
        \item  We propose a simple yet powerful and effective modification in the binarization process, by including a binary periodic function.
        \item We introduce a continuous, periodic sinusoidal function as a differentiable surrogate of the binary periodic function during the back-propagation process, suitable for stochastic gradient methods. 
        \item We mathematically analyze the quantization error of BiPer and show that it can be controlled by the frequency of the periodic function. This leverages a flexible initialization scheme for the binary weights, that balances the QE and network performance accuracy.      
	\item Experiments on benchmark data sets demonstrate the advantages of BiPer for the classification task with respect to state-of-the-art BNN approaches. BiPer outperforms prior works by up to $1\%$ on CIFAR-10, and $0.4\%$ on ImageNet.
\end{itemize}

%% file: sec/2_related_works.tex
\section{Related Work} 
\label{sec:related_works}

\subsection{Binary Neural Networks}
Among pioneering work on binary neural networks, BinaryConnect introduced a training scheme involving binary weights during training, and real-valued weights during the forward pass for weight updates \cite{BinConnect}. Subsequent approaches have focused on designing differentiable surrogates for the sign function to reduce the effect of the quantization error. For instance, a differentiable soft quantization (DSQ) approach was proposed in \cite{DSQ_2019}, which introduced a tanh-alike differentiable asymptotic function to estimate the forward and backward of the sign function. An improved BNN (BNN+) \cite{BNNplus} employs a SignSwish activation function to modify the back-propagation of the sign function and includes a regularization that encourages the weights around binary values. A more recent approach proposed an approximation of the sign function that varies along the training process \cite{AdaSTE2022}. 
The authors in \cite{RAD_2019} introduced an approach to training BNN by focusing on the regularization of activation distributions. Such regularization term encourages the binary activations to follow a balanced distribution during training. On the other hand, the rotated BNN \cite{NEURIPS2020_rotatedbnn} incorporates rotation operations into the training process, through a rotation-invariant loss function that mitigates the sensitivity of BNNs to input rotations. IR-Net \cite{Qin_2020_CVPR} employs a mechanism that preserves information in the form of real numbers during forward propagation and utilizes it in the backward pass. ReCU \cite{Recu2021} addresses the issue of dead weights in binary neural networks, by introducing a reconfiguration mechanism that revives inactive weights during training. ResTE \cite{ResTE_2023_ICCV} proposed an approach for training binary BNNs with a Rectified Straight Through Estimator, which addresses the limitations of the traditional STE. A rectification was introduced into the gradient estimation process during back-propagation, to enhance the accuracy of gradient signals, leading to improved training stability and convergence for BNNs. MST \cite{MST_2023_ICCV} proposed an approach for compressing and accelerating BNNs through the use of a Minimum Spanning Tree (MST). The MST-compression method leverages the structural characteristics of the network to identify an optimal subset of weights, so that  redundant connections are pruned, leading to a more compact BNN.

\subsection{Periodic Functions in Deep Neural Networks}
Periodic functions were successfully introduced as activations within neural network architectures in \cite{sitzmann2019siren}, where a sinusoidal function is used in a multilayer perceptron (MLP) to encode a spatial field into the network weights. In contrast to typical non-linear activation functions such as ReLU or tanh, periodic functions are capable of representing complex natural signals and their derivatives. For this reason, they have been since employed for 3D-aware image synthesis to represent scenes as view-consistent radiance fields\cite{Chan_2021_CVPR,Or-El_2022_CVPR}, as well as encoding model equations, like Partial Differential Equations (PDE), as a component of physics informed neural networks (PINN) \cite{pinn_periodicFunc}. It should be noted that all the aforementioned approaches use full-precision models, while this work employs a binary periodic function to promote binary weights in the forward model, and a sinusoidal function to better approximate the gradients of the binary weights in the backpropagation process. 
To the best of our knowledge, this is the first work that exploits the characteristics of periodic functions to improve the performance of BNNs.

%


\section{BNN Preliminaries}
\label{problem}
In this section, we introduce the essential mathematical formulations of neural network binarization. 
The main feature of BNN is to constrain the weights of a DNN as well as its activations to have binary values. The most common approach to obtain binary weights and activations consists of applying the sign function to a real-valued variable as
\begin{equation}
	w^q = \phi(w)=\text{Sign}(w) = \begin{cases}
		1,  & \text{if }w \geq  0, \\
		-1, & \text{otherwise},
	\end{cases}
 \label{eq:sign}
\end{equation}
where $w^q$ is the binarized variable and, $w$ is the real-valued variable. Here, we consider the binarization of a CNN model, however, these ideas can be extended to any neural model. Considering the binary weights $\mathbf{w}^q$ and the activations $\mathbf{a}^q$, the convolution operation can be formulated as 
\begin{equation}
	\label{eq:bin_conv}
	\mathbf{y} = \mathbf{w}^q \oplus \mathbf{a}^q,    
\end{equation}
where the symbol $\oplus$ denotes bit-wise operations including XNOR and BITCOUNT. It has been previously shown that the direct binarization of the convolution result in \eqref{eq:bin_conv} introduces large quantization errors. To alleviate this problem \cite{xnornet_2016} proposed to introduce real-valued scaling factors  $\alpha$ and $\beta$ to the weights and activations, respectively. Thus, the binary convolution can be expressed as 
\begin{align}
	\mathbf{y} = \alpha \beta \odot \left(\mathbf{w}^q \oplus \mathbf{a}^q \right),
\end{align}
where $\odot$ denotes element-wise multiplication. Further, the work in \cite{bulat2019xnor}, showed that combining $\alpha$ and $\beta$ in a single scaling factor $\gamma$ results in better performance. 
Since the gradient of the sign function is zero almost everywhere, the optimization of the weight parameters is incompatible with back-propagation algorithms based on gradient descent. To solve this problem, the work in \cite{BinConnect}, introduced the straight-through estimator (STE) to approximate the gradient with respect to weights as
\begin{equation}
\label{eq:STE}
	\frac{\partial\mathcal{L}}{\partial {w}}= \frac{\partial\mathcal{L}}{\partial {w}^q}\frac{\partial{w}^q}{\partial {w}}\approx \frac{\partial\mathcal{L}}{\partial {w}^q}. 
\end{equation}
More precisely STE uses the identity function as a differentiable surrogate of the sign function. It means that the gradient with the latent full-precision weights straightly equals to the gradient of the binarized outputs, which is also the origin of the name straight through estimator. To reduce the gradient error the clip function is also commonly employed, yielding to the approximation 
\begin{equation}
	\frac{\partial\mathcal{L}}{\partial {w}}= \frac{\partial\mathcal{L}}{\partial {w}^q}\frac{\partial{w}^q}{\partial {w}}\approx \frac{\partial\mathcal{L}}{\partial {w}^q} \mathbf{1}_{\mathcal{A}}(w),
\end{equation}
where $\mathbf{1}_{\mathcal{A}}$ is the indicator function of the clipping set $\mathcal{A}=\{w \in \mathbb{R}:|{w}|\leq 1\}$. While, in principle, the binarization of activations can be performed in the same manner as for the weights, several approaches have shown that using a different approximation for binarizing activations leads to improved performance \cite{DSQ_2019,liu2020bi,Qin_2020_CVPR}. In this work, following \cite{liu2020bi}, we use a piece-wise polynomial function to approximate the gradient with respect to the activations as follows

\begin{align}
    \frac{\partial \mathcal{L}}{\partial a}=\frac{\partial \mathcal{L}}{\partial a^q} \cdot \frac{\partial a^q}{\partial a} \approx \frac{\partial \mathcal{L}}{\partial a^q} \cdot \frac{\partial F\left(a\right)}{\partial a}    
\end{align}
with
\begin{align}
    \frac{\partial F\left(a\right)}{\partial a}= \begin{cases}2+2 a, & \text { if }-1 \leq a<0 \\ 2-2 a, & \text { if } 0 \leq a<1 \\ 0, & \text { otherwise }\end{cases}
\end{align}
where $a$ represents the real activation values, while $a^q$ represents the binary activations. The main limitation of the STE is that the gradient approximation introduces a considerable inconsistency between the forward and backward passes. To reduce the degree of inconsistency, several previous studies have explored alternative gradient estimators. 
Nevertheless, it has been shown that reducing the estimation error often results in highly divergent gradients, that harm the model training and increase the risk of gradient vanishing and gradient exploding \cite{ResTE_2023_ICCV}. 

%% file: sec/3_method.tex
\section{BiPer}
\label{sec:method}

\begin{figure}[H]
    \centering
    \includegraphics[width=0.98\linewidth]{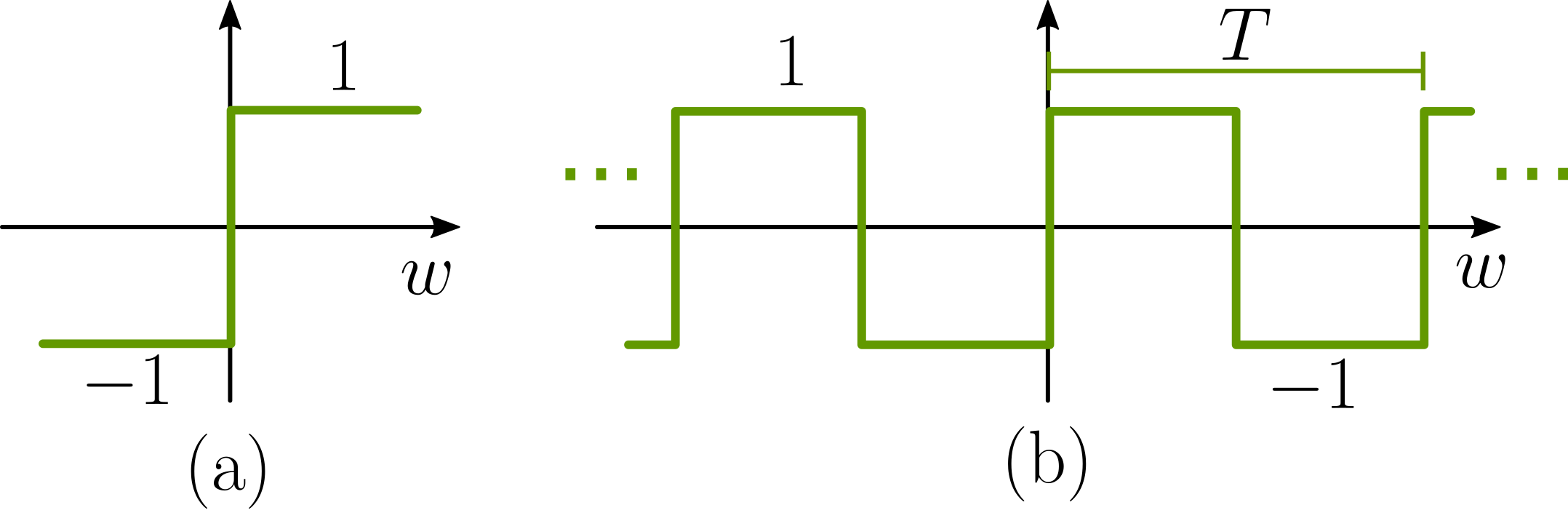}
	\caption{{(a) Sign function. (b) Binary periodic function.}}
	\label{fig:biper}
\end{figure}

To overcome the gradient and quantization error challenges from existing binarization methods and their gradient approximation functions, we propose to use a binary periodic function or square wave function (see Fig. \ref{fig:biper}) instead of just the sign function to model the binary weights. In contrast to the sign function described in section \ref{problem} and depicted in Fig. \ref{fig:biper}(a), which is always negative for negative values of $w$, the proposed periodic function (Fig. \ref{fig:biper}(b)) can reach positive and negative values in the whole domain of the latent weights.

It should be pointed out that the gradient of the periodic function still faces the problem of being zero almost everywhere, therefore, it cannot be directly integrated within a back-propagation algorithm based on gradient descent. To solve this problem, we first rewrite the square wave function as
	\begin{equation}
		w^q = \textrm{Sign} \left(\textrm{sin}(\omega_0 w)\right),
		\label{eq:biper} 
	\end{equation}
where $\omega_0=\frac{2\pi}{T}$ is the angular frequency. We note that this corresponds to applying the sign function to the first harmonic of the periodic function. Based on \eqref{eq:biper}, we can approximate the gradient with respect to the weights as
	\begin{equation}
        \label{eq:grad_biper}
		\frac{\partial\mathcal{L}}{\partial {w}}= \frac{\partial\mathcal{L}}{\partial {w}^q}\frac{\partial{w}^q}{\partial {w}}\approx \frac{\partial\mathcal{L}}{\partial {w}^q} \frac{\partial{\hat{w}}}{\partial {w}}
	\end{equation}
	where
	\begin{equation}
            \label{eq:sin_real}
		\hat{w} = \textrm{sin}(\omega_0 w).
	\end{equation}
 Note that the last differential term in \eqref{eq:grad_biper} corresponds to the gradient of a continuous differentiable sinusoidal function, which is also a smooth periodic function, and proportional to the frequency $\omega_0$. Comparing Eq. \eqref{eq:grad_biper} to the STE approximation in \eqref{eq:STE}, it can be seen that the proposed approximation is the product of the STE estimator of a periodic function and a proportional constant given by the frequency. In the following section, we explore this property to control the quantization error with the frequency $\omega_0$.

\subsection{Quantization Error Analysis}
\label{sec:QEAnalysis}
This section shows that an additional advantage of using the periodic function
is its flexibility to control the quantization error. 
In particular, we mathematically demonstrate how a lower quantization error can be achieved by setting the fundamental period of the wave function. To this end, let us first assume that the latent weights roughly follow the zero-mean Laplace distribution, i.e., $\mathcal{W} \sim La(0,b)$  \cite{baskin2021uniq,zhang2021differentiable, banner2019post}. Since the weights $\hat{w}$ in \eqref{eq:sin_real} before quantization are a function of a random variable, they are also a random variable $\mathcal{\hat{W}} \in [-1,1]$. 
Computing the probability density function (pdf) of a random variable $\mathcal{Y}=g(\mathcal{X})$ from the pdf of $\mathcal{X}$ $\left(f_{\mathcal{X}}(x)\right)$ can be easily done employing the method of transformation \cite{pishro2014introduction}, if the function $g$ is differentiable and strictly increasing or decreasing, i.e., strictly monotonic. Thus, the pdf of $\mathcal{Y}$ can be computed as 
\begin{align}
f_\mathcal{Y}(y)= \begin{cases}
\frac{f_\mathcal{X}\left(x_1\right)}{\left|g^{\prime}\left(x_1\right)\right|}=f_X\left(x_1\right) \cdot\left|\frac{d x_1}{d y}\right| & \text { where } g\left(x_1\right)=y \\ 
0 & \text { if } g(x)\neq y. 
\end{cases} 
\nonumber
\end{align}
The more general case in which $g$ is not monotonic requires splitting the domain into $n$ intervals, so that $g$ is strictly monotonic and differentiable on each partition. Then, the pdf can be obtained as
\begin{align}
\label{eq:pdf_y}
f_\mathcal{Y}(y)=\sum_{k=1}^n \frac{f_\mathcal{X}\left(x_k\right)}{\left|g^{\prime}\left(x_k\right)\right|}=\sum_{k=1}^n f_\mathcal{X}\left(x_k\right) \cdot\left|\frac{d x_k}{d y}\right|,    
\end{align}
where $x_1, \cdots, x_n$ are real solutions to $g(x)=y$. For BiPer, since the periodic function from Eq. \eqref{eq:sin_real} is not monotonic, we can use \eqref{eq:pdf_y} to compute the pdf of $\hat{\mathcal{W}}$ using the pdf of $\mathcal{W}$. Letting $f_\mathcal{W}(w)= \frac{1}{b}exp(|w|/b)$ denote the pdf of $\mathcal{W}$, and setting $g$ as the sine function, we can divide the sinusoidal function into subsequent intervals of $T/2$ where it is strictly increasing or decreasing, alternately. For simplicity, let us assume $\omega_0=1$, so that $T=2\pi$. Then, the solutions $w_k$ to $\text{sin}(w)=\hat{w}$ are given by
\begin{align}
    w_k = (-1)^k \text{arcsin}(\hat{w}) + \pi k,
\end{align}
for some $k \in \mathbb{Z}$. Considering that the latent weights follow a Laplace distribution, using $w_k$ and Eq. \eqref{eq:pdf_y}, we obtain the pdf of the weights before binarization as 
\vspace{-1em}
\small
\begin{align}
\label{eq:sum_infty}
f_{\hat{\mathcal{W}}}(\hat{w})=\sum_{k=-\infty}^\infty \frac{1}{2 b } \frac{1}{\sqrt{1-\hat{w}^2}} \exp \left(\frac{-|(-1)^k \text{arcsin} (\hat{w}) + \pi k|}{b }\right).   
\end{align} 
\normalsize
As it is shown in the supplementary material, the summation in \eqref{eq:sum_infty} converges to the probability density function of the latent weights before binarization $\hat{w}$ for an arbitrary frequency $\omega_0$ given by
\small
\begin{align}
	f_{\hat{\mathcal{W}}}(\hat{w}) & =\frac{1}{2 b \omega_0} \frac{1}{\sqrt{1-\hat{w}^2}} \exp \left(\frac{-|\arcsin (\hat{w})|}{b \omega_0}\right)\nonumber \\
	& +\frac{1}{2 b \omega_0} \frac{1}{\sqrt{1-\hat{w}^2}} \cosh \left(\frac{\arcsin (\hat{w})}{b \omega_0}\right) \frac{1}{e^{\pi / b \omega_0}-1}.
\end{align}
\normalsize
\begin{figure}[H]
	\centering
	\includegraphics[width=0.75\linewidth]{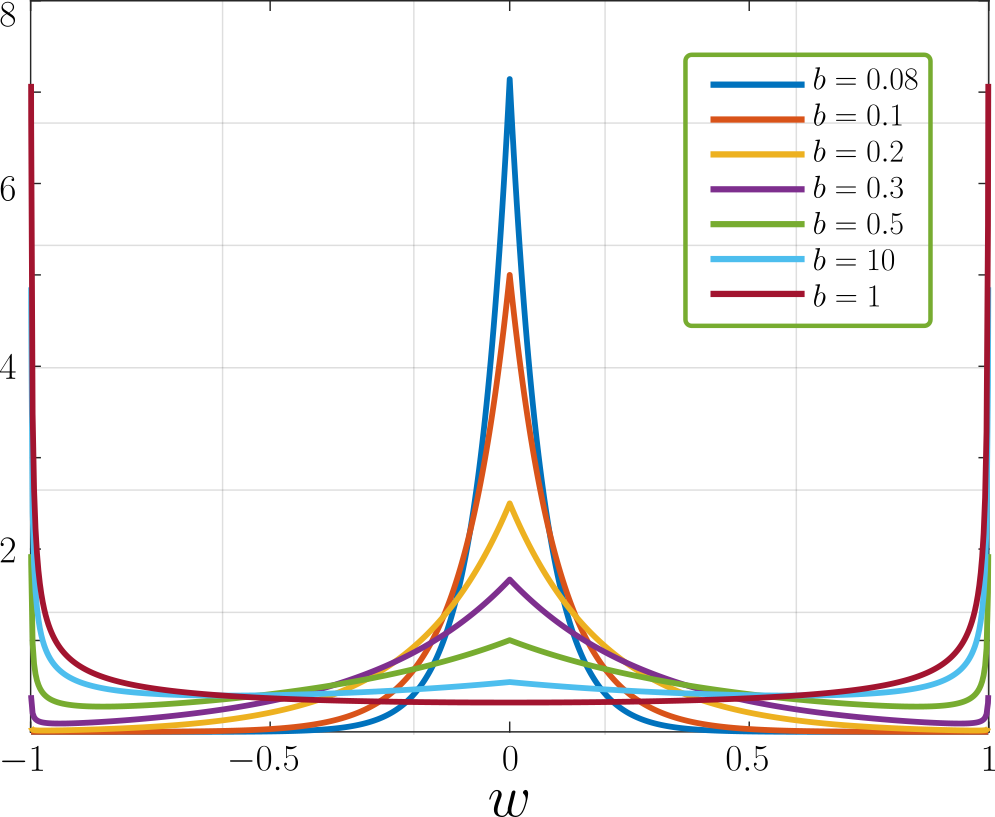}
	\caption{Probability density function of $\hat{w}=\textrm{sin}(\omega_0 w)$ assuming that the random variable $w$ follows a Laplace distribution with parameter $b$ and a fixed value of $\omega_0=1$.}
	\label{fig:pdf}
\end{figure}

\begin{figure}[H]
	\centering
	\includegraphics[width=0.8\linewidth]{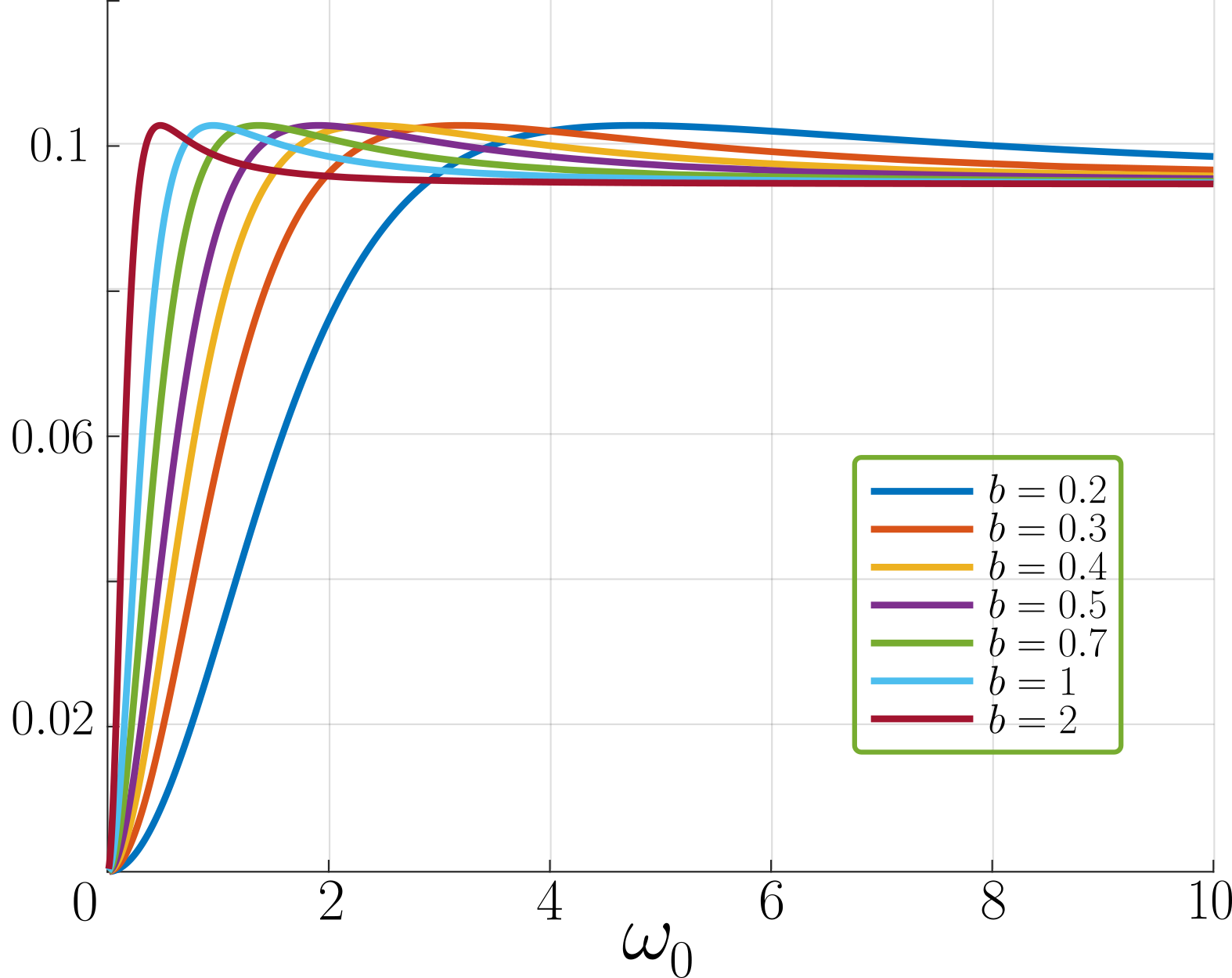}
	\caption{Quantization error as a function of the frequency $\omega_0$ for different values of $b$. The proposed BiPer approach is able to control QE with the frequency of the periodic function.}
	\label{fig:QE}
\end{figure}

Figure \ref{fig:pdf} depicts the distribution of the weights for different values of the Laplace distribution parameter $b$ and a fixed frequency $\omega_0=1$. Note that different from the random variable $\mathcal{W}$ which can take any real value, the codomain of the random variable $\hat{\mathcal{W}}$ is $[-1,1]$. From Fig. \ref{fig:pdf} we can observe that when the value of $b$ increases the pdf of $\hat{w}$ behaves as an arcsin distribution with values concentrated around -1 and 1. This reduces the quantization error in comparison to the Laplacian distribution. Also, a similar behavior occurs when the frequency value increases for a fixed $b$. To further analyze these observations consider the QE defined as
\begin{align}
\label{eq:QE1}
	\mathrm{QE} = \int_{-\infty}^{+\infty} f_\mathcal{W}\left(w\right)\left(\textrm{sin}(\omega_0 w)-\gamma \operatorname{sign}\left(\textrm{sin}(\omega_0 w)\right)\right)^2 \mathrm{~d} w,
\end{align}
where $f_\mathcal{W}$ is the density distribution function of the latent weights. Using the fact that $|x| = x \textrm{Sign}(x)$ along with the properties of the absolute value, we can rewrite Eq. \eqref{eq:QE1} as
\begin{align}
	\mathrm{QE} = \int_{0}^{+\infty} \frac{1}{b}  \exp \left(\frac{-w}{b}\right)\left(|\textrm{sin}(\omega_0 w)|-\gamma\right)^2 \mathrm{~d} w.
\end{align}
The solution to this integral is given by (see supplementary material for more details on this calculation) 

\begin{align}
\label{eq:QE}
	\mathrm { QE }=\frac{2(\omega_0 b)^2}{4(\omega_0 b)^2+1}-\frac{2 \gamma \omega_0 b\left(e^{\pi / \omega_0 b}+1\right)}{\left.(\omega_0 b)^2+1\right)\left(e^{\pi / \omega_0 b}-1\right)}+\gamma^2. 
\end{align}
On the other hand, the optimal solution of the scaling factor $\gamma$ in \eqref{eq:QE1} can be computed as
\begin{align}
\label{eq:gamma}
	\gamma = \mathbb{E}\{|\text{sin}(\omega_0 w)|\}= \frac{ \omega_0 b\left(e^{\pi / \omega_0 b}+1\right)}{\left.(\omega_0 b)^2+1\right)\left(e^{\pi / \omega_0 b}-1\right)}.
\end{align}
Replacing $\gamma$ from \eqref{eq:gamma} into Eq. \eqref{eq:QE}, we can rewrite the QE as a function of the frequency $\omega_0$ and the parameter $b$. Figure \ref{fig:QE} illustrates the QE as a function of the frequenct $\omega_0$ for different values of $b$. It can be seen that the maximum QE is $0.102835$, which occurs when the product $b\omega_0\approx0.954882$. This means that for a fixed value of $b$ the maximum QE occurs at $\omega_0 \approx 0.954882/b$.
Furthermore, from this value, we note that by increasing the frequency to $\infty$, the QE converges to $0.5 - \dfrac{4}{\pi^2}$. Nonetheless, increasing $\omega_0$ poses an additional issue since the gradient is proportional to the frequency value, thus it will diverge. This result is consistent with recent literature, which states that minimizing the QE yields to divergent gradients  \cite{ResTE_2023_ICCV}. On the other hand, 
from Fig. \ref{fig:QE}, we can also see that the quantization error can be reduced by decreasing the frequency value. However, this corresponds to the trivial case when  $\gamma \textrm{Sign}(\textrm{sin}(\omega_0) w)$ goes to zero as well as the weights.
Clearly, this is not a practical approach either since there would not be information available to update the gradients. 

It is worth noting that in contrast to current approaches that progressively reduce the QE to zero, BiPer does not meet this QE value. Nonetheless, further explorations can adapt state-of-the-art surrogate estimators to smoothly converge from the sine function to the square wave. These studies however, fall beyond the scope of this paper, which explores an alternative binarization technique.


%% file: sec/4_experiments.tex
\section{Experiments}
\label{sec:experiments}
We comprehensively evaluated BiPer for image classification, with widely used neural network architectures, i.e., ResNet and VGG-Small, trained on benchmark datasets (CIFAR-10 and ImageNet). In the following, we first describe the experiments setup. 
Then, we present ablation experiments conducted on CIFAR-10. Finally, we compare BiPer with state-of-the-art BNN approaches in terms of performance and complexity on both data sets.

\subsection{Experiments Setup}
\subsubsection{Data Sets}
\paragraph{CIFAR-10 \cite{cifar-10dataset}:} 
Consists of 60,000 images of size $32\times32$, divided into 10 categories. Training set contains 50,000 images 
and the remaining 10,000 are used for testing.
\vspace{-0.3cm}
\paragraph{ImageNet \cite{Imagenet_ILSVRC12}:} A challenging data set because of its larger size and more diverse image categories. Among its multiple versions, we adopted the widely used ILSVRC12 version, which is divided into 1,000 categories, from which 1.2 million are training images and, 50,000 test images. ImageNet is the most widely used data set to report results on binary networks and, it allows us to show for the first time that binary networks can perform competitively on a large-scale data set.
\subsubsection{Network Architectures}
On CIFAR-10, we evaluated BiPer using ResNet-18/20 and VGG-Small.  For ResNet-18/20, we adopted the double skip connections as in \cite{liu2020bi}, to provide fair comparisons. On ImageNet (ILSVRC-2012) we chose to binarize ResNet-18/34. Following \cite{bethge2019training}, the downsampling layers are not quantized, and the double skip connections \cite{liu2020bi} were included. Following standard procedures of the comparison methods, we binarized all layers but the first and last.
\subsubsection{Training Details and Procedures}
BiPer approach was implemented in PyTorch because of its flexibility and powerful automatic differentiation mechanisms. Our models were executed on a NVIDIA RTX 3090 GPU.
All experiments used stochastic gradient descent (SGD) optimization with $0.9$ momentum. We followed the data augmentation strategies in \cite{he2016deep}, which
include random crops and horizontal flips.\vspace{-1em}
\paragraph{Two stage training:} 
Recent works have shown that an appropriate initialization is often required to improve network performance. BNN initialization strategies employ an adaptation of high precision Neural Networks (NNs) \cite{glorot2010understanding,he2016deep}, or start from a pre-trained real-valued NN. 
To alleviate feature quantization adverse effects, two-stage training strategies are generally employed \cite{martinez2019training,bulat2019improved}. Particularly, in the first stage the network is trained with real weights and binary features. 
Then, in the second stage, a warm weight initialization is employed based on the binary representation of the output weights from the first stage, and the model is fully trained to binarize the weights. Thus, the problem splits into two sub-problems: weight and feature binarization. 
In BiPer, we propose a two-stage training where, the first stage uses real-valued weights $\hat{w}$ as in Eq. \eqref{eq:sin_real}, and the second stage uses the weight binarization from Eq.\eqref{eq:biper}. 
%
%
If not otherwise specified, the weight-decay was fixed at $5e-4$ for the first stage and $5e-5$ for the second stage. 
The learning rate was set to $0.1$ in the first stage and, $0.01$ in the second stage. In both stages the learning rate was adjusted by the cosine scheduler.
 

\subsection{Ablation Studies}
To investigate the actual contributions of BiPer training stages, we conducted ablation studies on CIFAR-10. In all these experiments, ResNet-18/20 was used as the backbone.
\subsubsection{Impact of Frequency $\omega_0$}
\begin{figure*}[ht!]
    \centering
    \begin{subfigure}[t]{0.3\textwidth}
        \centering
        \includegraphics[height=1.25in]{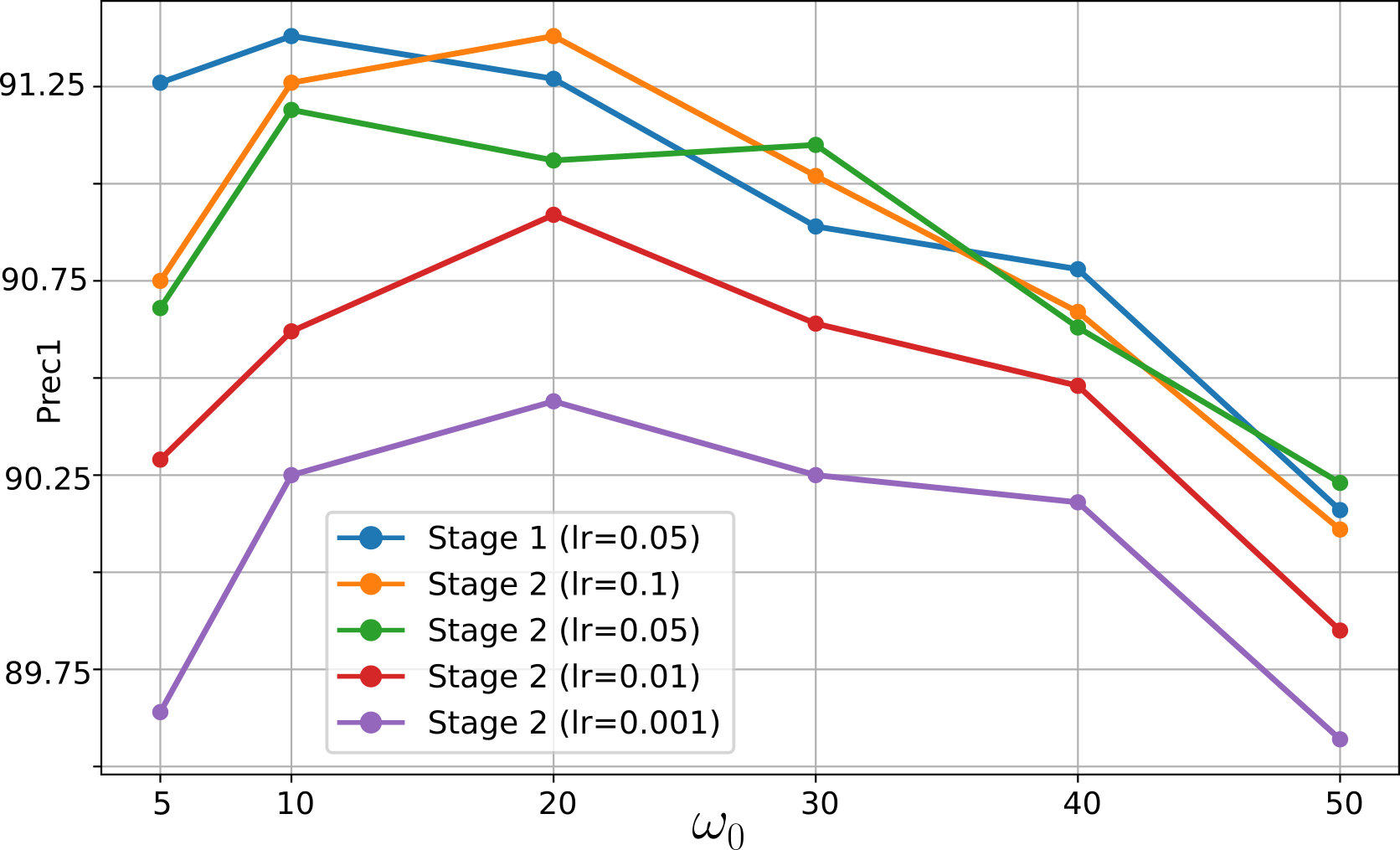}
        \caption{$\omega_0$ \vs Top-1 Precision}
    \end{subfigure}%
    ~ 
    \begin{subfigure}[t]{0.3\textwidth}
        \centering
        \includegraphics[height=1.25in]{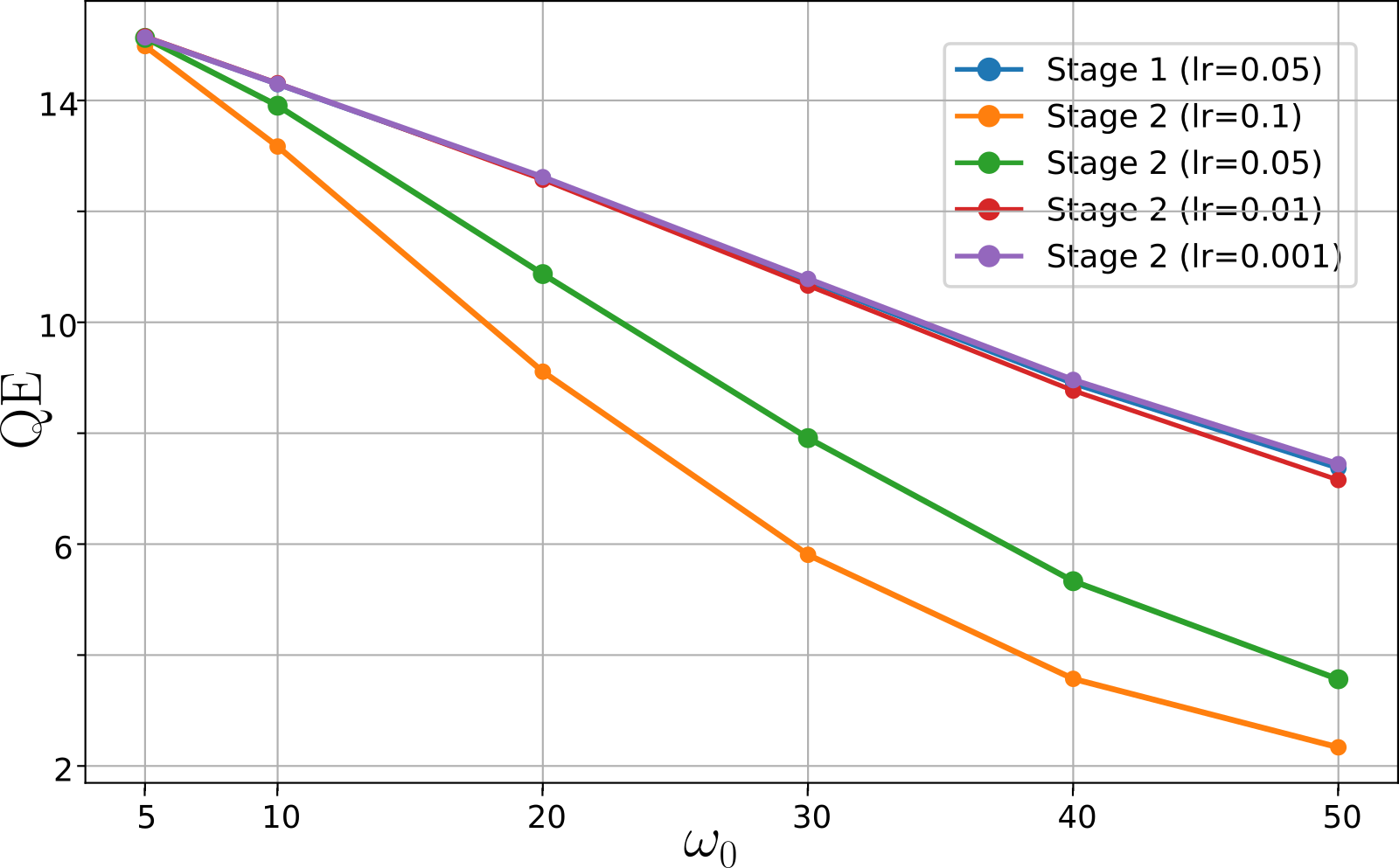}
        \caption{$\omega_0$ \vs $\textrm{QE}$}
    \end{subfigure}
    ~ 
    \begin{subfigure}[t]{0.3\textwidth}
        \centering
        \includegraphics[height=1.25in]{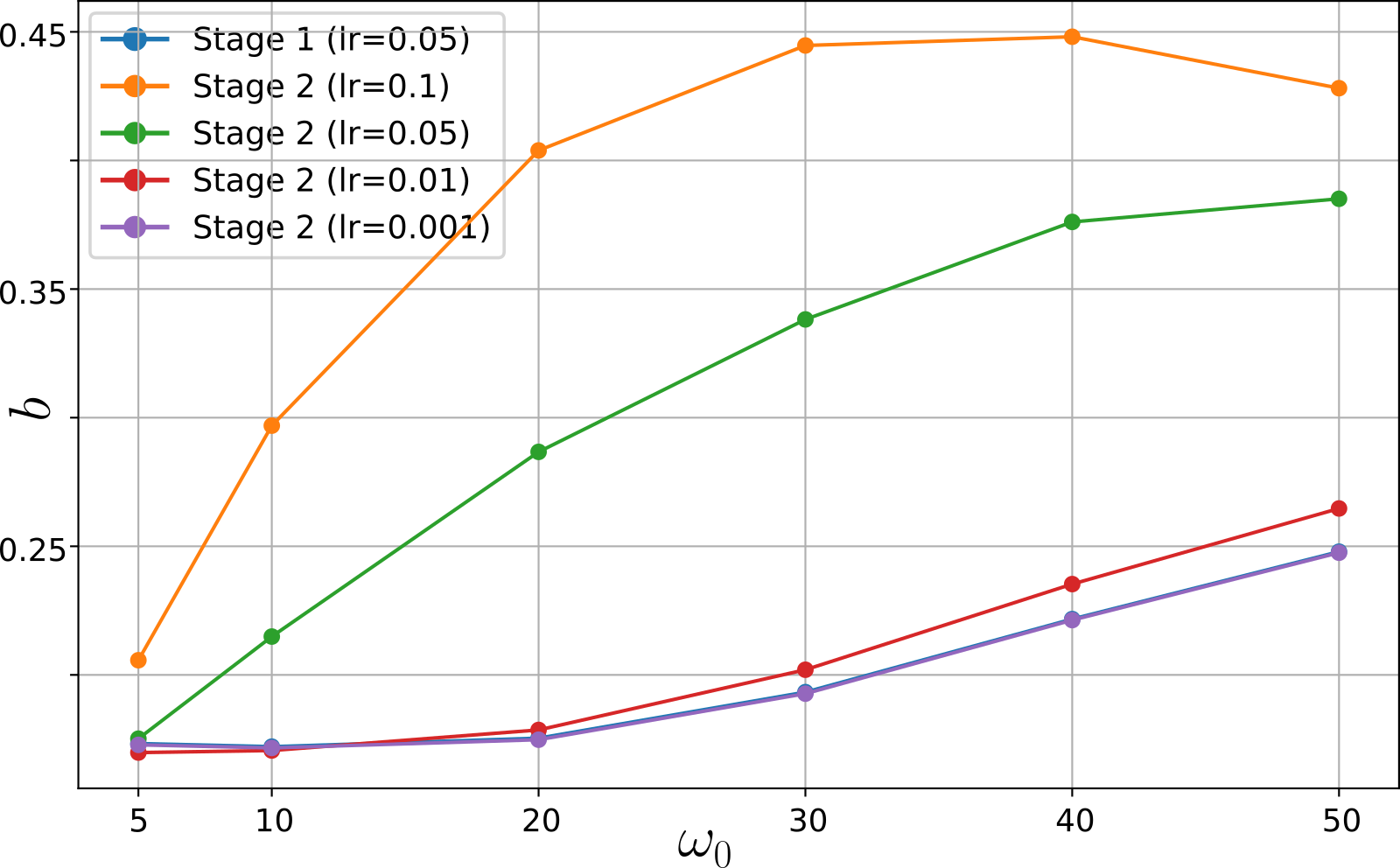}
        \caption{$\omega_0$ \vs $b$}
    \end{subfigure}
    \caption{Impact of the frequency of the periodic function $\omega_0$ on BiPer (a) Top-1 classification precision, (b) Quantization error and, (c) Weight distribution parameter $b$, for the CIFAR-10 data set. }
    \label{fig:freqablation}
\end{figure*}

BiPer introduces the frequency of the periodic function $(\omega_0)$ as a relevant hyperparameter. Section \ref{sec:QEAnalysis} showed that $\omega_0$ can control QE, and particularly, the weight QE depends on the product of $\omega_0$ and the parameter $b$. Thus, for a given value of $b$, increasing the frequency implies lower QE. To verify these theoretical results, we trained BiPer Stage 1 for $200$ epochs, with a lr$=0.05$, weight decay $5e-4$, and varying $\omega_0$. 
Figure \ref{fig:freqablation} illustrates the obtained stage 1 results for (a) Top-1 validation precision, (b) QE and (c) maximum likelihood estimated parameter $b$. 
These results validate our theoretical findings. 
However, consistently with the state of the art, reducing the QE degrades network performance. It is worth noting that during the first stage, there is no QE for the weights and, the values reported in Table 1 correspond to the QE obtained from weight binarization at the end of stage 1 training. This same value corresponds to the QE of the initialization for stage 2. 

We further verified the effect of $\omega_0$ on the full BiPer model. To this end, we trained BiPer Stage 2 for different learning rates, using the outputs of Stage 1 as warm start. 
The results are also presented in Figure \ref{fig:freqablation}. 
Interestingly, Fig. \ref{fig:freqablation} (a) shows that the best Stage 1 model ($\omega_0=10$) does not result in the best performance of the full binary model. Furthermore, ensuring the lowest QE in Stage 1, 
does not result in the best network performance either. For the frequency hyperparameter, we found that an intermediate frequency $\omega_0= 20$ balances the initial QE and precision of the full binary model. Moreover,  in some cases the final BiPer model achieves higher precision than the initial model trained with real weights, e.g. $\omega_0=20,30$. 
To the best of our knowledge, this is the first empirical analysis of the impact of the quantization error in the  initialization. 

A third experiment was conducted in order to verify the contribution of the periodic function in the binarization process. In particular, we are interested on evaluating whether BiPer improvements on classification precision are due to the use of the periodic function instead of the two-stage training methodology. Therefore, we trained BiPer following the proposed two-stage training, but we just employ the sign function to binarize the weights and activations as in \eqref{eq:sign}.
Table \ref{table:two-stage-real} summarizes the results for this experiment. We can see that stage 1 attained higher precision and QE compared to stage 2. Specifically, precision degradation reached $0.9\%$ and QE decreased approximately $0.2$. This behavior is similar to that from the case when $\omega_0=5$ in Fig. \ref{fig:freqablation}, where we obtained high precision and high QE on stage 1, but in the second stage, the performance drops.
This occurs because lower frequencies promote small argument values for the sinusoidal function, and $\textrm{sin}(x)\approx x$ for small values of $x$. 
\begin{table}[h!]
	\caption{BiPer two-stage training without using a periodic function in the binarization. \vspace{-1em}}
	\label{table:two-stage-real}
	\begin{center}
		\begin{tabular}{c|c|c|c}
                \hline
			 
			\multicolumn{2}{c|}{Stage 1}    & \multicolumn{2}{c}{Stage 2}\\ \hline \hline
			Precision & QE         & Precision  & QE      \\ 
			   93.93  & 0.2351      & 93.01     & 0.0584      \\ 
            \hline
		\end{tabular}
	\end{center}
\end{table}
\vspace{-3em}

\subsubsection{Binary Weights - Real Activations}
This experiment considers only weight binarization using the proposed BiPer approach. Thus, we trained a resnet-20 for the CIFAR-10 data set, keeping real activation values.  Quantitative results are reported in Table \ref{table:only_w}, where it can be noted that the proposed approach has competitive performance with the state-of-the-art approach LCR-BNN \cite{shang2022lipschitz}. Furthermore, it is important to highlight that the degradation accuracy compared to a full precision model is less than $1\%$ 

\begin{table}[h!]
	\caption{Performance comparison for binary weights and real activations in BiPer, compared to state-of-the-art methods on CIFAR-10. W/A: bit length of weights and activations. FP: Full precision model.}\vspace{-1em}
	\label{table:only_w}
	\begin{center}
		\begin{tabular}{llll}
			\hline
			Network                       & Method & W/A & Top-1 \\ \hline \hline
			\multirow{6}{*}{ResNet-20}    & FP        & 32/32     & $91.7\%$       \\ 
			& DoReFa        & 1/32     & $90.0\%$      \\ 
			& LQ-Net       & 1/32    & $90.1\%$      \\ 
			& DSQ    & 1/32     & $90.2\%$      \\
			& IR-Net       &  1/32    & $90.8\%$      \\
                & \textbf{LCR-BNN}       &  1/32    & {$\textbf{91.20}\%$}      \\
			& \textbf{BiPer (Ours)}       &  1/32   & $\textbf{91.20}\%$ 
            \\\hline
		\end{tabular}
	\end{center}
 \vspace{-2em}
\end{table}


\subsection{Comparison with SOTA methods}
\subsubsection{Experiments on CIFAR-10}

For ResNet-18, we compared the proposed BiPer approach with existing methods such as RAD\cite{RAD_2019}, IR-Net \cite{Qin_2020_CVPR}, LCR-BNN \cite{shang2022lipschitz}, RBNN \cite{lin2020rotated}, ReCU \cite{Recu2021}, ReSTE \cite{ResTE_2023_ICCV}, MST \cite{MST_2023_ICCV}, and DIR-Net \cite{qin2023distribution}. For VGG-Small, we compared BiPer with the XNOR-Net \cite{xnornet_2016}, BNN \cite{BNN_2016}, DoReFa \cite{zhou2016dorefa}, IR-Net \cite{Qin_2020_CVPR}, RBNN \cite{lin2020rotated}, DSQ, SLB \cite{yang2020searching}, ReCU \cite{Recu2021} and ReSTE \cite{ResTE_2023_ICCV}. The classification precision results are shown in Table \ref{table:CIFAR10}, and it can be seen that BiPer provides the best performance for ResNet-18 and ResNet-20, while it provides competitive results for the VGG-small network. Specifically, the proposed approach obtains a Top-1 accuracy of $93.75\%$ when binarizing ResNet-18, improving classification precision in almost $1\%$ against ReCU and $0.55\%$ against the recent Minimum Spanning Tree (MST) strategy. Furthermore, the accuracy loss against the full precision model is reduced to just $1\%$. Recall that BiPer is capable of providing these results without requiring additional network modules or regularization functions.

\begin{table}[]
	\caption{BiPer performance comparison with  state-of-the-art BNN on CIFAR-10. W/A: bit length of weights and activations. FP: full precision model.}\vspace{-1em}
	\label{table:CIFAR10}
	\begin{center}
		\begin{tabular}{llll}
			\hline
			Network                       & Method & W/A & Top-1 \\ \hline \hline
			\multirow{6}{*}{ResNet-18}    & FP        & 32/32     & $94.8\%$       \\ 
			& RAD        & 1/1     & $90.5\%$      \\ 
			& IR-Net       & 1/1    & $91.5\%$      \\ 
                & LCR-BNN & 1/1 & $91.80 \%$ \\
			& RBNN    & 1/1     & $92.2\%$      \\
			& ReCU       &  1/1    & $92.8\%$      \\
                & ReSTE     & 1/1 & $92.63 \%$ \\
                & MST & 1/1 & $93.2 \%$ \\
                & DIR-Net & 1/1 & $92.8 \%$ \\
			& \textbf{BiPer (Ours)}       &  1/1   & $\textbf{93.75}\%$       \\\hline
			\multirow{6}{*}{ResNet-20}    & FP        & 32/32     & $92.1\%$       \\ 
			& DoReFa        & 1/1     & $79.3\%$      \\ 
			& DSQ       & 1/1    & $84.1\%$      \\ 
			& SLB    & 1/1     & $85.5\%$      \\
			& IR-Net       &  1/1    & $86.5\%$      \\
			& {ReCU}       &  1/1   & ${87.4}\%$ \\                
                & \textbf{BiPer (Ours)}       &  1/1   & $\textbf{87.5}\%$ 
            \\\hline
			\multirow{9}{*}{VGG-small}    & FP        & 32/32     &   $94.1\%$     \\
			& XNOR-Net       & 1/1     & $89.8\%$      \\
			& BNN        & 1/1     &  $89.9\%$    \\
			& DoReFa        & 1/1     &  $90.2\%$    \\
			& IR-Net        & 1/1    & $90.4\%$      \\
			& RBNN        & 1/1    & $91.3\%$      \\
			& DSQ        & 1/1     & $91.7\%$     \\
			& SLB        & 1/1     & $92.0\%$     \\
			& \textbf{ReCU}        & 1/1    & $\textbf{92.2}\%$     \\                
			& {BiPer (Ours)} & 1/1    & $92.11\%$ \\\hline 
		\end{tabular}
	\end{center}
 \vspace{-2em}
\end{table}

\subsubsection{Experiments on ImageNet}

\begin{table}[]
	\caption{BiPer performance comparison with  state-of-the-art BNN on ImageNet. W/A: bit length of weights and activations. FP: full precision model.}\vspace{-1em}
	\label{table:imagenet}
	\begin{center}
		\begin{tabular}{lllll}
			\hline
			Network                       & Method & W/A & Top-1 & Top-5\\ \hline \hline
			\multirow{12}{*}{ResNet-18}    & FP        & 32/32     & $69.6\%$ & $89.2\%$       \\ 
			& XNOR-Net        & 1/1     & $51.2\%$   & $73.2\%$   \\ 
                & Bi-Real Net        & 1/1     & $56.4\%$   & $79.5\%$   \\
                & PCNN        & 1/1     & $57.3\%$   & $80.0\%$   \\
			& IR-Net       & 1/1    & $58.1\%$  & $80.0\%$    \\
   
                & BONN        & 1/1     & $59.3\%$   & $81.6\%$   \\
                & LCR-BNN        & 1/1     & $59.6\%$   & $81.6\%$   \\
                & HWGQ        & 1/1     & $59.6\%$   & $82.2\%$   \\
                & RBNN    & 1/1     & $59.9\%$    & $81.9\%$  \\
                & FDA       &  1/1    & $60.2\%$  & $82.3\%$    \\
			& ReSTE       &  1/1    & $60.88\%$  & $82.59\%$    \\   
                & ReCU       &  1/1    & $61.0\%$  & $82.6\%$    \\
                & DIR-Net       &  1/1    & $60.4\%$  & $81.9\%$    \\
			& \textbf{BiPer (Ours)}    &  1/1   & $\textbf{61.4}\%$ & $\textbf{83.14}\%$  \\\hline
			\multirow{7}{*}{ResNet-34}    & FP        & 32/32     & $73.3\%$ & $91.3\%$       \\ 
                & Bi-Real Net        & 1/1     & $62.2\%$   & $83.9\%$   \\
			& IR-Net       & 1/1    & $62.9\%$  & $84.1\%$    \\
                & RBNN    & 1/1     & $63.1\%$    & $84.4\%$  \\
			& ReSTE       &  1/1    & $65.05\%$  & $85.78\%$    \\   
                & ReCU       &  1/1    & $65.1\%$  & $85.8\%$    \\
                & DIR-Net       &  1/1    & $64.1\%$  & $85.3\%$    \\
			& \textbf{BiPer (Ours)}    &  1/1   & $\textbf{65.73}\%$ & $\textbf{86.39}\%$     
			\\\hline 
		\end{tabular}
	\end{center}
 \vspace{-2em}
\end{table}

We also evaluate the proposed BiPer approach using ResNet-18 and ResNet-34, and training on the large-scale ImageNet dataset. Table \ref{table:imagenet} shows a number of SOTA quantization methods over ResNet-18 and ResNet-34, including XNOR-Net \cite{xnornet_2016}, Bi-Real Net \cite{liu2020bi}, PCNN \cite{gu2019projection}, IR-Net \cite{Qin_2020_CVPR}, BONN \cite{gu2019bayesian}, LCR-BNN \cite{shang2022lipschitz}, HWGQ \cite{cai2017deep}, RBNN \cite{lin2020rotated}, FDA \cite{xu2021learning}, ReSTE \cite{ResTE_2023_ICCV}, ReCU \cite{Recu2021}, and DIR-Net \cite{qin2023distribution}. We can observe that the proposed BiPer approach in the 1W/1A setting achieves the best Top-1 and top-5 accuracy for both network architectures. Specifically, for ResNet-18, we attained a top-1 validation accuracy of $61.4\%$, outperforming the second-best result of $61.0\%$ achieved by ReCu. Furthermore, our top-5 performance reached $83.14\%$, surpassing the second-best result of $82.6\%$, also achieved by ReCU. Likewise, for ResNet-34, we achieved the highest top-1 and top-5 accuracies, namely $65.73\%$ and $86.39\%$, respectively. These results improve the second-best method (ReCU) by $0.63\%$ and $0.59\%$ in top-1 and Top-5 accuracies, respectively. 
The extensive comparison results presented in Table \ref{table:CIFAR10} and Table \ref{table:imagenet} demonstrate the effectiveness of BiPer on classification tasks. 





%% file: sec/5_conclusion.tex
\section{Conclusions}
\label{sec:conclusion}
An approach for neural network binarization using a binary periodic function or square wave, dubbed BiPer, has been proposed. 
To improve gradient stability we employed a sinusoidal function with the same period of the square wave as a differentiable surrogate during the backward pass.
This simple, yet powerful modification tackles the problem of standard gradient mismatch between forward and backward steps during network training, providing a suitable alternative that can be incorporated within back-propagation algorithms based on stochastic gradient descent. 
Mathematical analysis of BiPer quantization error demonstrated that it can be controlled by the frequency of the periodic function. This leverages a flexible initialization scheme for the binary weights, that balances the QE and network performance accuracy.
The advantages of BiPer for the classification task were demonstrated through experimental results for two benchmark data sets, i.e., CIFAR-10 and ImageNet. Comparisons with respect to state-of-the-art BNN approaches showed that BiPer outperforms prior works by up to $1\%$ on CIFAR-10, and up to $0.63\%$ on Imagenet, respectively.
Although this work tested the BiPer approach for classification, it can be easily extended to other high-level tasks without increasing the complexity. Thus, 
BiPer opens new horizons for neural network quantization by analyzing the frequency of the periodic function and its relation to the network quantization error. In this regard, we have shown that the QE can be controlled by the frequency.
These advancements collectively contribute to the maturation of BNNs as viable alternatives for real-world applications, offering a compelling trade-off between model size, computational efficiency, and accuracy.\\
\noindent \textbf{Acknowledgment.} This work was supported by the Vicerrector\'ia de Investigación y Extensión of UIS, Colombia under the research project 3735.
